\documentclass{article}

\usepackage{arxiv}

\usepackage[utf8]{inputenc} 
\usepackage[T1]{fontenc}    
\usepackage{hyperref}       
\usepackage{url}            
\usepackage{booktabs}       
\usepackage{amsfonts}       
\usepackage{nicefrac}       
\usepackage{microtype}      
\usepackage{lipsum}
\usepackage{graphicx}

\title{A Linked Aggregate Code for Processing Faces\\(Revised Version)} 

\author{
    Michael J. Lyons\\
    Human Information Processing Laboratories\\
    Advanced Telecommunications Research International\\
    Kyoto, Japan\\
   \And
    Kazunori Morikawa\\
    Laboratory of Vision Research\\
    Rutgers University\\
    Piscataway NJ USA\\
}

\begin{document}
\maketitle
\begin{abstract}
\baselineskip=24pt 
  A model of face representation, inspired by the biology of the visual system, is compared to experimental data on the perception of facial similarity.  The face representation model uses aggregate primary visual cortex (V1) cell responses topographically linked to a grid covering the face, allowing comparison of shape and texture at corresponding points in two facial images. When a set of relatively similar faces was used as stimuli, this ``linked aggregate code'' (LAC) predicted human performance in similarity judgment experiments. When faces of perceivable categories were used, dimensions such as apparent `sex' and `race' emerged from the LAC model without training.  The dimensional structure of the LAC similarity measure for the mixed category task displayed some psychologically plausible features but also highlighted differences between the model and the human similarity judgements.  The human judgements exhibited a racial perceptual bias that was not shared by the LAC model. The results suggest that the LAC based similarity measure may offer a fertile starting point for further modelling studies of face representation in higher visual areas, including studies of the development of biases in face perception. 
\end{abstract}
{\let\thefootnote\relax\footnote{Michael Lyons is currently with Ritsumeikan University and Kazunori Morikawa is currently with Osaka University.}}
\newpage
\baselineskip=24pt
\section{Introduction}
It is well established that higher visual areas, such as
inferotemporal cortex (area IT), contain populations of neurons
specialized for processing facial stimuli (Bruce {\it et al.} 1981;
Perrett {\it et al.} 1982).  The mechanism by which these areas
process faces, however, is not well characterized.  It is not known,
for example, how the output of areas early in the visual pathway is
transformed by higher-level areas such as the face-selective areas of
IT and a neurobiologically realistic model of face processing is
impossible at the moment.  We do know, however, that face and object
visual information passes through the primary visual cortex (area V1)
on the way to the form processing areas in the inferotemporal cortex and
some aspects of the processing in these earlier visual areas have been
characterized. Hence it is interesting to examine the transformation
of facial images by this early stage of processing and explore what
aspects of human facial perception can be accounted for by 
this first step towards a neurobiological model of face processing.

In this paper, we study a facial representation built from
topographically ordered aggregates of the spatial frequency and
orientation-selective neural responses characteristic of the primary
visual cortex (area V1) (De Valois and De Valois, 1988). In addition
to the neurobiological rationale, this representation is motivated by
the observation that unlike general objects, faces cannot be well
distinguished by a parts-level description such as the
Recognition-by-components theory (Biederman, 1987), which places
emphasis on the list of component parts of a complex object. However,
the component features of individual faces are largely invariant: the
parts are always of the same class. Rather it is variances in the
spatial relations of parts and textural differences that must be coded
to distinguish two faces.  V1 neurons are retinotopically organized,
have relatively small receptive fields, and are sensitive to texture
and shape on several spatial scales. Our aim in this paper is to start
from V1 receptive field properties and build towards a face
representation model that captures some aspects of human face
perception. We emphasize that the claim is not that faces are represented
in terms of the direct output of V1 neurons, but rather that it may be fruitful to explore the strengths and limitations of such a model.

A third motivation for this study is that V1 inspired visual
representations have already found useful application in computer
vision (e.g. Daugman 1985) as well as a popular approach to automatic
face representation as explored by the Malsburg group 
(Lades {\it et al.} 1993).

In this work, we investigated the predictive power and the
limitations of a model of face representation based on the output of
V1 neurons.  Other more widely studied aspects of facial information
processing such as facial memory, sex classification, or facial
expression recognition are expected to be beyond the scope of the
the simple filter-based model examined here, however all of these
phenomena are dependent upon some notion of physical similarity.  All
models of such processes necessarily involve computation of the
the similarity between two faces or an equivalent measure of
relatedness. If two faces appear similar to human observers, then in a
good input representation they should be situated close to each other.

The similarity between faces is an interesting phenomenon in its own right. Human faces are distinguishable by relatively subtle differences in the shapes and relative positions of features. Nonetheless, faces can appear quite different. On the other hand, we often perceive a resemblance between faces that can be difficult to describe verbally. 
What exactly makes two faces look similar or different is not well
understood.  The phenomenon of perceived facial similarity and its underlying neural mechanism has received relatively little attention
compared with the amount of research on face recognition and
classification (Bruce {\it et al.} 1987; Bruce {\it et al.} 1993;
Hill {\it et al.} 1995; Gray {\it et al.} 1995; O'Toole {\it et al.} 1991).

Preliminary findings from this project were presented in a talk at the ARVO'96 conference (Lyons and Morikawa 1996). A similar approach has been applied to a study of facial expression perception (Lyons {\it et al.} 1997, Lyons {\it et al.} 1998a).

\section{The Model}
The model we test in this paper starts from two rather general
assumptions: (1) visual information passes through area V1 and is
output by the complex cells of V1 on the way to higher cortical
areas and, (2) a degree of facial correspondence is necessary
to compare two faces.

Comparison of faces requires access to detailed information about
relatively minor differences in the shapes and textures of the faces. Such
information can be coded by the cells of the primary visual
cortex that are sensitive to spatial patterns at multiple
orientations and spatial scales and which preserve the topographic
organization of visual space in their physical arrangement in the
cortical sheet. A general statement of the hypothesis studied in this
work is that a useful representation for faces may be obtained by
linking the outputs of multi-scale and orientation filters to a structure having the topography of the face.  This allows comparison of spatial patterns at corresponding positions, scales and orientations. We term this class of face representations ``Linked Aggregate Codes'' (LAC) because they link localized pattern descriptors to the topography of the face. The
particular LAC face representation used in the experiments in this paper is similar to the face code used by a successful face recognition algorithm developed by the Malsburg group (Lades {\it et al.} 1993)  but with filter properties derived from data on the visual system.

Our study is concerned with a computational stage at which position
and size constancy is already achieved. The face-selective cells of IT
have low sensitivity to translation and scale changes of the stimulus.
Malsburg's ``dynamic link architecture'' (Malsburg 1981) for achieving
correspondence between image and model domains is one example of a
biological model for the process of ``linking'' image features in a
topographic representation of an object or face. However, our
conclusions are not dependent on the validity of the ``dynamic link
architecture'' model or any other model of image-model matching. Instead, the present work is concerned with the
representation of facial shape and assumes only that approximate
correspondence of facial images exists. Our conclusions are
independent of the details of how this correspondence is obtained.

The LAC similarity measure used in this work is based on a simple model
of the primary visual cortex (Fig. 1). The spatial pattern of pixel values
centred on image position $\vec{r}$ is represented by a population
response vector, $\bf{R}$.  $\bf{R}$ is calculated by convolving the
image, $I(\vec{r})$, with a set of Gabor filters, $G_{\vec{k},+}$ and
$G_{\vec{k},-}$, which approximate V1 simple cells having even and odd
receptive fields respectively and spatial frequency and orientation
tuning determined by the wave-vector index, $\vec{k}$ 
(Marcelja 1980; Daugman 1980, 1985; Jones and Palmer 1987):
\begin{equation}
R_{\vec{k},\pm} ( \vec{r}_{0} ) = \int G_{\vec{k},\pm} (\vec{r}_{0}, \vec{r})
I(\vec{r}) d \vec{r},
\end{equation}

\begin{equation}
G_{\vec{k},+}(\vec{r}) = k^2 (e^{-k^2 \|\vec{r} - \vec{r}_{0}\|^2/2 \sigma ^2}
cos(\vec{k} \cdot (\vec{r} - \vec{r}_{0}))) - e^{-\sigma ^2 /2}),
\end{equation}

\begin{equation}
G_{\vec{k},-}(\vec{r})  = k^2 e^{-k^2 \|\vec{r} - \vec{r}_{0}\|^2/2 \sigma ^2}
sin(\vec{k} \cdot (\vec{r} - \vec{r}_{0})).
\end{equation}
\begin{figure}[t]
\begin{center}
\includegraphics[width=.8\linewidth]{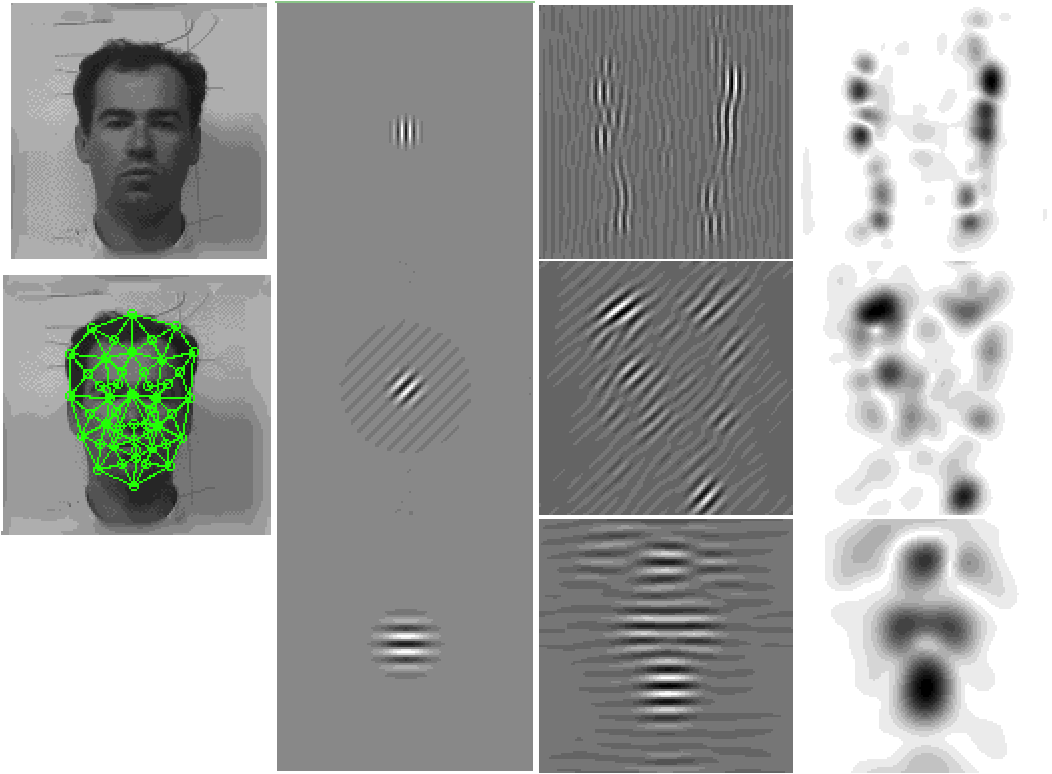}
\caption{Representation of a facial image with Gabor filters
at several spatial frequencies and orientations. At the extreme left
of the figure are shown the original facial image of size 128x128
pixels and 256 gray levels and a grid used to sample the Gabor transform.  We convolved the image with 36 Gabor filters (3 spatial frequencies and 6 orientations at 2 phases). The figure shows three sample filters in the leftmost column of the center block.  The center column shows the real part of the response of the image to the Gabor transform, while the rightmost column shows the response amplitude.  Hence, each point of the image generates 18 numbers corresponding to the response amplitude at each spatial position.  These 18 numbers constitute the components of the Gabor vector, which serves as an approximate model of the response of a bank of cells in the primary visual cortex.}
\label{Fig1}
\end{center}
\end{figure}
In these equations, the multiplicative factor $k^2$ compensates for
the $k^{-2}$ dependence of the power spectrum of natural images so
that each Gabor vector component has similar magnitude (Field 1987)
In equation (2) the integral of the cosine Gabor filter,
\begin{equation}
e^{- \sigma^2 / 2}
\end{equation}
is subtracted from the filter to render it insensitive to
the absolute level of illumination.  The sine filter does not depend
on the absolute illumination level.  Three spatial frequencies were
used having the wave-numbers:
\begin{equation}
k = \{\frac{\pi}{2},\frac{\pi}{4},\frac{\pi}{8} \}
\end{equation}
as measured in inverse pixels. The highest frequency was set at half the Nyquist sampling
frequency. The frequency levels are spaced at octaves; $\sigma = \pi $
was used in all calculations, giving a filter bandwidth of about an
octave, independent of the frequency level. Six wave-vector
orientations were used, equally spaced at intervals of $30 \deg$ from
$0 \deg$ to $180 \deg$. These parameters approximate values found in
neurophysiological data (Daugman 1985; Jones and Palmer 1987).  With
the viewing geometry of the experimental setup, the three frequency
levels are $\{ 4.9, 2.5, 1.2 \}$ cycles/degree (cpd), well within the
range of the peak sensitivity of human vision (0.8 - 16 cpd, with
maximum sensitivity at about 5 cpd) (for example see Wilson and Gelb
1984).  Inclusion of an additional lower frequency level did not
improve the results. The images used do not have sufficient resolution
to calculate the next octave higher frequency in the Gabor vector
representation.

The components of the Gabor vector, $R_{\vec{k}}$, are defined as the
amplitude of the combined even and odd filter responses:
\begin{equation}
R_{\vec{k}} = \sqrt{R^{2}_{\vec{k},+}+R^{2}_{\vec{k},-}} 
\end{equation}
The response amplitude is
less sensitive to position changes than are the individual filters and
approximates the output of non-end-stopped complex cells in V1
(for example see DeValois and DeValois 1988).

To compute facial similarity, responses of cells having the same
spatial frequency and orientation preference are compared at
corresponding points in the two facial images. The normalized dot
product is used to quantify the similarity of two Gabor response
vectors. The LAC similarity of two facial images is the average of the
Gabor vector similarity over all corresponding facial points. Since
Gabor vectors at neighbouring pixels are highly correlated and
redundant, it is sufficient to calculate the average on a sparse grid
covering the face (Fig. 1).  It is probable that some facial locations
and filters of some spatial frequencies and orientations at certain
facial positions are more salient for processing faces than others.
As a first-order approximation, we adopt the hypothesis that all
filters are equally weighted. Determining differential weighting of
the filters will be investigated in a later study.

In this study, facial graphs were positioned by hand or by using an
automatic graph matching algorithm (Lades {\it et al.} 1993). Only approximate correspondence is required by this model: the similarity function depends on the amplitude of the Gabor transform which is less sensitive to translation and small scale changes than the phase-sensitive linear filters.

\section{Experiments}
In the present study, we conducted two experiments to compare this
model and human judgments of facial similarity and showed that the LAC
similarity correlates well with perceived similarity, can capture
psychologically salient dimensions of faces and produce a similarity
space resembling that of human observers.

To check that a significant correlation of the model with human
perception does not obtain with any arbitrary model of
facial similarity, we also calculated a control similarity measure. We consider a control measure derived from the pixel
information in the facial image. We formed pixel vectors with
components taken as pixel values from square patches centred at each
grid point, using the same grid as used for the LAC model. We calculated the pixel-based similarity using the Euclidean distance between pixel
vectors averaged over all corresponding nodes in two facial
graphs. The pixel patches we tried ranged in sizes from 5x5 to 20x20.
The use of the pixel values is mathematically equivalent to a principal
component analysis (PCA) representation that retains all principal
components because the PCA representation is just a
rotation of coordinates (Duda \& Hart 1973). However, an optimal PCA
representation would differentially weight the components as with
the more recently proposed ``Fisherface'' representation (Belhumeur
{\it et al.} 1997).  As stated at the beginning of this section, our goal was not to find the similarity measure which agrees most with human perception.
\subsection{Experiment 1}
\subsubsection{Methods}
Experiment 1 studied facial similarity within a single category. We
used ten gray-scale frontal photographs of European origin (EO) male faces as
stimuli.  Each image was 384x384 pixels with the background included.  , Each face subtended approximately 6.5$^\circ$ of visual angle in a computer screen, viewed from a distance of about 60 cm.  The
faces used were relatively similar to one another, and none of them
had facial hair or eyeglasses.  Thirty-two undergraduates (10 American
and 22 Japanese) participated as subjects. None of them were
aware of the purpose of the experiment.  On each trial, each subject was
shown three faces on a computer screen, one in the top half of the
screen, two in the bottom half.

The experiment required subjects to choose which of the two bottom faces looked more similar to the top face.  We asked the subjects to make judgments based on the overall shape and configuration of facial features and to ignore the hairstyle, facial expressions, and overall lightness and darkness of faces.  A single session used all possible combinations of three faces except those triads for which the two bottom faces were identical.  These conditions result in 450 trials per subject presented in random order, including the 90 for which the target and one of the bottom faces were the same.  These triads serve to monitor subjects' attentiveness, but do not count in the subsequent analysis. This number is close to the maximum that a subject can perform in a single session. Therefore, the stimulus set is limited to about ten faces. Despite this limitation, we used the triad forced-choice paradigm since it offers the subject a facial similarity judgment task that is unequivocal.
\subsubsection{Results}
One of the main conclusions of this paper is that when the LAC model, a
simple measure of physical similarity between two faces, has
considerable predictive power when the task is to compare two faces
within a single facial category. However, the model begins to show its
limitations for a more complex task: comparison of faces between two apparent categories. However, even for this more complex task, the model exhibited a natural clustering of psychologically
distinguishable types of face.

The analysis calculated the concordance between subjects as the percentage of triads for which subjects agreed out of 360 trials.  The LAC similarity measures were computed for the 360 triads. Whichever bottom face produced a higher similarity is taken to be the model's response.  The concordance between subjects and the model was calculated using the same procedure.  The mean concordance between the LAC measure and subjects was $64.18\%$, which was somewhat higher than the mean concordance of $59.94\%$ for all possible 496 subject pairs. The difference between the two was marginally significant (p $<$ .056).\footnote{Because the distribution of concordances is unknown, we used a distribution-free bootstrap procedure to estimate the standard error and significance level (Efron and Tibshirami, 1993). The estimated standard errors of the mean human-human concordance, the mean LAC-human concordance, and the difference between these two were 2.10\%, 0.84\%, and 2.24\%, respectively.} If the LAC similarity were unrelated to perceived similarity, the LAC-human concordance would be lower than the human-human concordance.  These results show that the LAC measure captures substantial aspects of facial similarity as perceived by humans, at least within the apparent facial category,  ``EO male.''

To give a more concrete notion of these results, Table 1 lists the
similarity rankings for each stimulus used in the experiment. The
table shows that agreement between the LAC measure and the data ranges over the full distribution of similarities. For example, out of the three faces that were rated most similar to each target face by
human subjects, 66.7\% were also chosen to be among the three most
similar faces by the LAC model.  Likewise, 70.0\% of the three least
similar faces as rated by human subjects were also among the three
least similar faces as rated by the LAC model.
\begin{figure*}[t]
\begin{center}
\includegraphics[width=.8\linewidth]{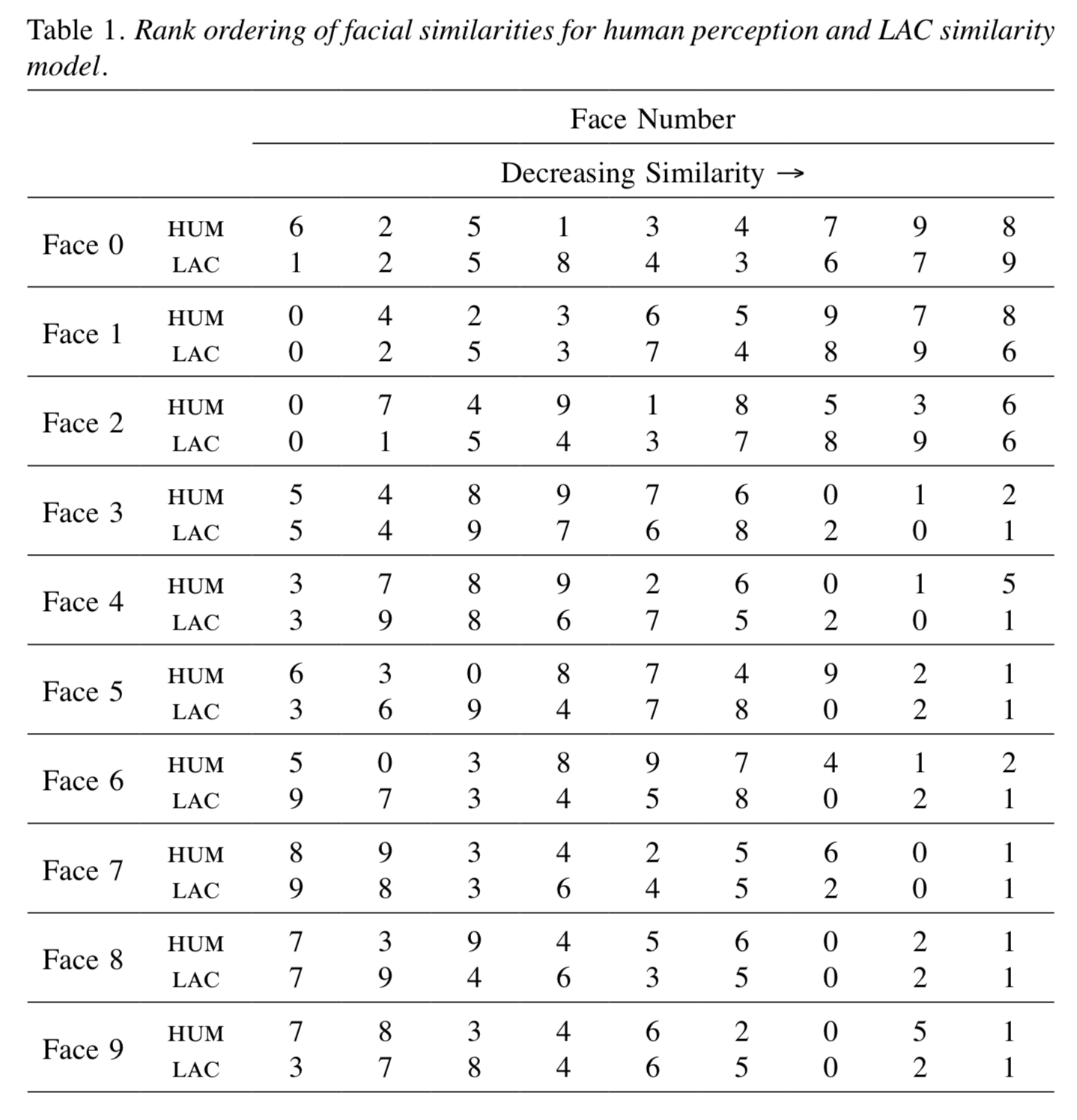}
\end{center}
\end{figure*}
As an additional comparison between model and data, we
averaged human response on the triad comparison task to generate a
similarity index between each pair of faces that we correlated with
the LAC measure. As there are no adjustable parameters in the LAC
similarity model, absolute values of the similarity measure are of
less interest than relative values, hence the model and data were
correlated. Spearman's rank correlation between the data-derived
similarity index and the similarity measure derived from the LAC model
was found to be $0.71$.

The average Spearman rank correlation coefficient of the pixel-patch
derived similarity measure with the data was $0.52$. Hence the Gabor
derived similarity model explains approximately twice as much of the
variation in the human data as the pixel-based control.

The subject pool is mixed because of the availability of subjects at our
labs in Japan and the United States and not chosen to study cross-cultural effects. The data in experiment 1 showed no significant cross-cultural biases.  Inter-subject concordance was $54.49\%$ for the Japanese subjects and $61\%$ for the Americans. Model-data concordance was $63.84\%$ for Japanese subjects and $64.9\%$ for the Americans. Spearman's rank correlation
coefficient of the averaged rating between the American subject pool and the Japanese subject pool is $0.85$. This level of correlation confirms the robustness of our basic experimental approach for measuring facial similarity since the two subject pools ran the experiment in different laboratories.

\subsection{Experiment 2}
\subsubsection{Methods}
Experiment 2 studied facial similarity between four facial appearance categories.  The stimuli used consisted of 16 gray-scale photographs of faces chosen from the following apparent categories: 4 European origin (EO) males, 4 EO females, 4 East Asian (EA) males, and 4 EA females.  Because this is too many faces for a triad similarity judgment task, we instead chose a pairwise similarity level judgment paradigm. The number of stimuli is the usable limit for a single subject judging of
similarities between faces from four categories.  The size of the
photographs was the same as that used in Experiment 1.  Eight American
undergraduates participated as subjects.  None of them was aware of
the purpose of the experiment.  On each trial, two faces were
presented side by side on a computer screen.  The task was to rate
the similarity between the two faces using a 10-point scale.  We asked subjects to base ratings on the overall shape and configuration of facial features and to ignore the hairstyle, facial expressions, and overall lightness or darkness of faces.  All possible combinations of two
different faces were used (120 pairs). Each experimental session
consisted of 3 blocks, each block comprising all pairs.  The order of
presentation within each block was randomized.  The first block was
considered as practice and excluded from the subsequent analysis.  
The experiments counterbalanced the left-right positions of faces across the second and third blocks.
\subsubsection{Results}
Similarity ratings on the second and third blocks were averaged and
then normalized within each subject, resulting in a 16x15 triangular
similarity matrix.  The overall correlation between the model and
data similarity values was 0.343---lower than what was found for
the within-category experiment. Eight individual matrices were averaged and subjected to non-metric multidimensional scaling (nMDS) (the ALSCAL algorithm was used, Takane {\it et al.} 1977).  Fig. 2 shows a two dimensional projection of the 3-D solution ($Stress = .096, R^2 = .910$).
\begin{figure}[t]
\begin{center}
\includegraphics[width=.8\linewidth]{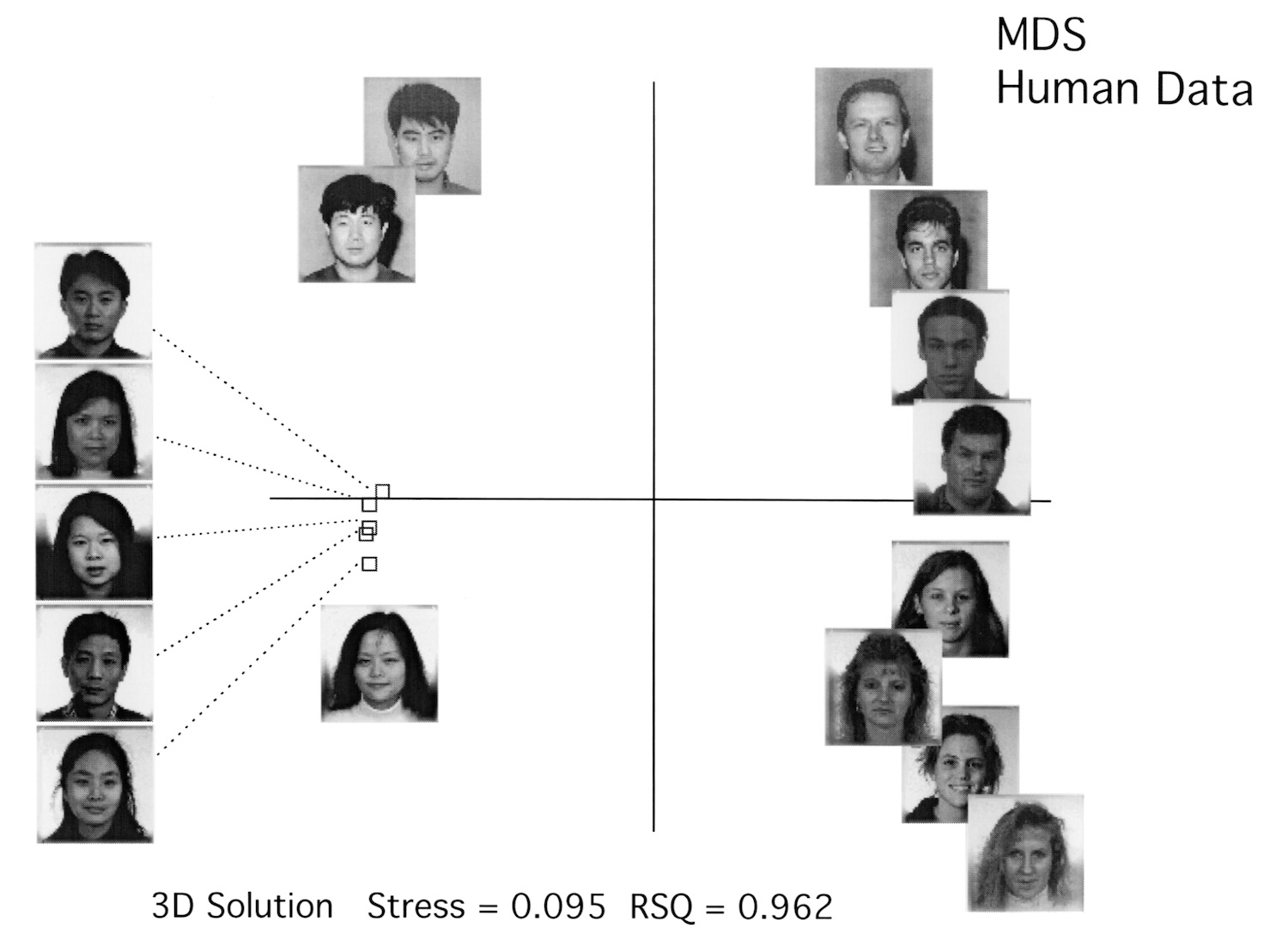}
\caption{The similarity space of human perception from
Experiment 2.  This is a 2-D projection of the 3-D nMDS solution.  The
vertical axis corresponds to perceived `sex,' the horizontal axis to perceived `race.' The
stress, a parameter representing badness of fit (Takane {\it et al.} 1977), was
0.095 (minimum value is 0). $R_{sq}$, a parameter representing
goodness of fit (Takane {\it et al.} 1977) was 0.962 (maximum value is 1).}
\label{Fig2}
\end{center}
\end{figure}
The two dimensions recovered represent perceived `race'\footnote{The term `race' is used in a limited sense in this paper to indicate faces distinguishable by an appearance that is perceptibly EA or EO. More generally, we recognize that `race' is not based on definable physical or biological differences but is a social construction, or identity, based on perceived differences. Likewise, we use the term `sex' here to distinguish appearance perceived as being `male' and `female' according to social conventions. We fully acknowledge the existence of human gender identities that are not described by these two apparent categories. A more detailed analysis of this topic is beyond the scope of the current article.} and perceived `sex,' thus demonstrating that perceived facial similarity reflects apparent facial categories.  The configuration of the EA faces includes a `sex' misclassification and is compressed compared to the EO faces. These results recall the ``other race'' psychological effect (Shapiro and Penrod 1986; Bothwell {\it et al.} 1989; Valentine 1991; O'Toole {\it et al.} 1994). The experimental subjects in this experiment were American, none of whom was of EA heritage. The detailed investigation of cross-cultural facial appearance perception between-categories, however, is beyond the scope of the current study.

The LAC similarity was also computed for the same 120 pairs of the 16 faces and the resulting similarity matrix was subjected to the same nMDS analysis.  Fig. 3 shows a two dimensional projection of the 3-D solution ($Stress = .095, R^2 = .962$).
\begin{figure}[t]
\begin{center}
\includegraphics[width=.8\linewidth]{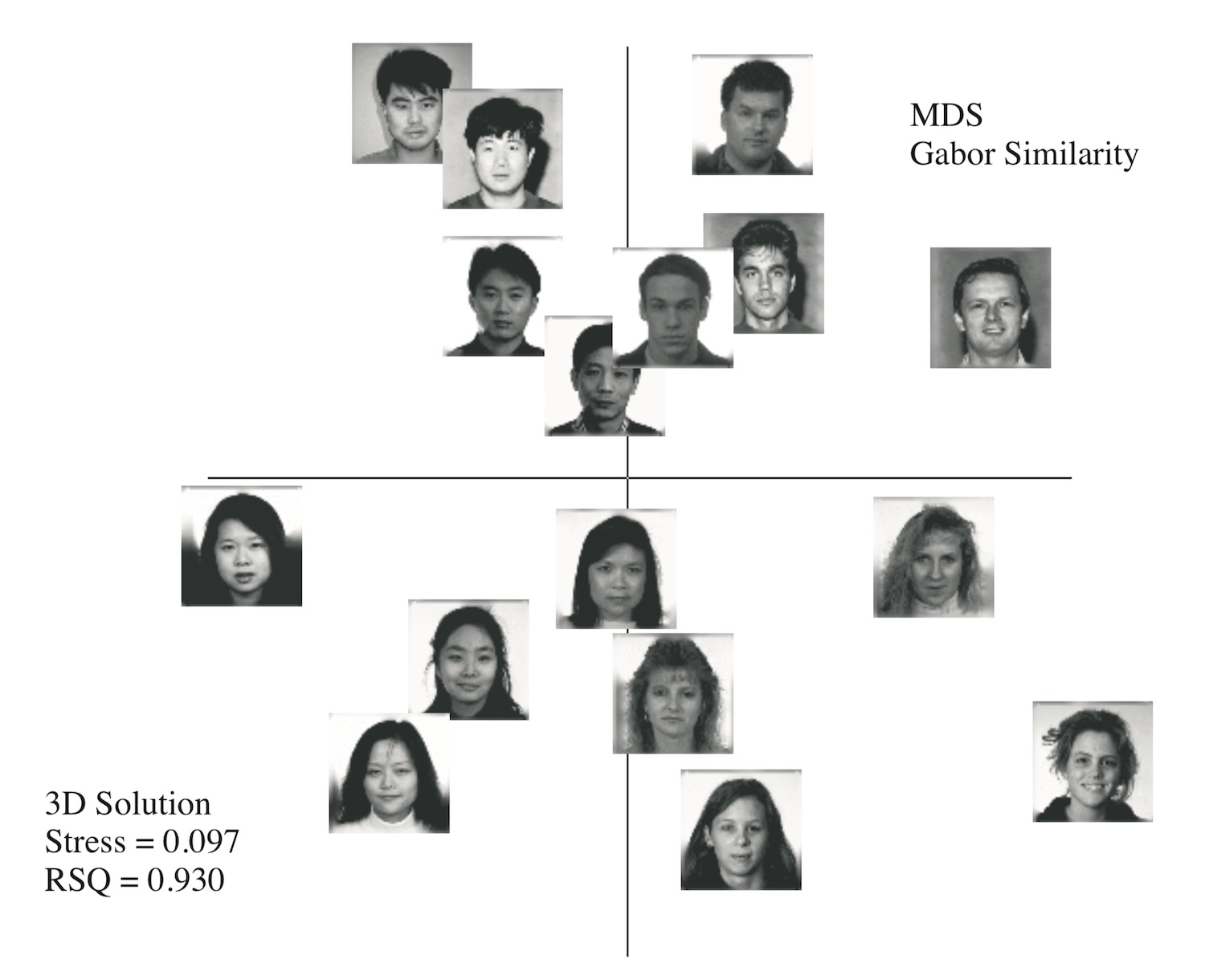}
\caption{The similarity space obtained from the Gabor filter
output for the stimuli used in Experiment 2.  This is a 2-D projection
of the 3-D nMDS solution.  The vertical axis corresponds to the apparent `sex,' the
horizontal axis to the apparent `race.' Stress = 0.097, $R_{sq}$ = 0.930.}
\label{Fig3}
\end{center}
\end{figure}
As with human perception, the two principle dimensions represent ostensible `race' and `sex' properties, and these are nearly orthogonal.  By contrast, the individual faces are more evenly distributed than for the space derived from human judgements. In other words, both the EA and EO faces are positioned more gradually on both axes. There is no tight clustering of facial categories as was seen with the human similarity judgement data. Overall, while the overall dimensional structure of the facial similarity space, as represented by the model, matched that as perceived by humans, the precise positioning of faces within the space differed.

The purpose of the nMDS analysis presented in this section is to
illustrate that the LAC similarity measure naturally recovers
psychologically meaningful clusters of facial appearance. However, as
shown by the correlation of similarity values, detailed agreement
between model and data for the mixed-category similarities is not as
high as when individuals of a single facial category are being
compared. Moreover, qualitative differences between the positioning of
faces in the two spaces are easily visible from a comparison of Figs. 2
and 3.
\section{Discussion}
Previous computational studies have adopted several different
approaches to representing faces.  One general approach to facial
coding employs the length of certain facial features and
inter-feature distances which are explicitly measured and used as data
for subsequent analysis such as discriminant function analysis (Burton {\it et al.} 1993). 
Such codes isolate the explicit geometrical shape information and discard information about facial
texture and shading.  A second general approach codes luminance values of pixels that are input to, for example, a principal components analysis (PCA) or to a back-propagation network trained to
recognize or classify faces by supervised learning (Gray {\it et al.} 1995; O'Toole {\it et al.} 1991).

Both pixel-based and geometry-based approaches have yielded valuable insights into face processing, but neither approach is motivated by the known biology of the human visual system. Indeed, the first stage of face coding by the primary visual cortex simple and complex cells uses information intermediate between explicit geometry, requiring highly spatially localized features, and the pixel-derived approaches, such as PCA methods, that usually involve receptive fields distributed over the entire input region. Gabor filters are local in both space and spatial frequency (Daugman 1985), allowing an integrated description of geometry and texture. LAC similarity implicitly codes facial measurements in the responses of filters with spatially distributed receptive fields, which is more plausible than an explicit representation of precise inter-feature distances.  The use of the amplitude of the filter response makes the LAC measure robust to small variations in the filter position on the face and small variations in the viewing angle.

For the single category experiment, using the concordance measure (the
percentage of agreed trials) we found a significant level of agreement between LAC similarity and human perception, which was at least as good as the inter-subject agreement. Direct correlations of
similarity values derived from the LAC similarity model and human
responses were also high, considering that there are no adjustable
parameters in the similarity model. 

An nMDS analysis was attempted for the model-derived within-category
similarity data, however, the fitting criteria indicated unsatisfactory fit for the number of dimensions tried (2-4).  Higher-dimensional solutions are of limited interest, given the small number of stimuli used for the within-category study. A previous experimental study (Rhodes 1988), described a 3 dimensional MDS solution for facial similarity perception (within the single category of EO male faces) and correlated the recovered dimensions with intermediate-level features derived from the spatial relations describing the dimensions of facial parts (e.g. mouth size, nose length) and configural dimensions (e.g. distance between the eyes).  However, the LAC representation is both lower-level, in that it consists only of non-linear spatial filter responses, and high-dimensional in that the input vector has several hundred dimensions which are a function not
only of the shape but the texture of the face.  The introduction of a
further dimensionality-reducing step, as a bridge between the filter responses and the type of features used by Rhodes, might
allow us to recover a lower-dimensional within-category space.  This
would be an interesting direction in which to extend the present
research. However, as noted above a larger number of faces would be
needed for such a study.  Alternate experimental paradigms for
measuring similarity data with a larger number of items based on tree
methods were employed in the study of Rhodes, cited above.

The fact that a low-dimensional solution obtained for the mixed-category experiment suggests that the similarity variation due to differences between-category dominated that due to differences between individuals.  While the mixed-category experiment demonstrated an explanatory power of the model, it also highlighted some of the shortcomings. The same qualitative dimensions emerged in the nMDS analysis of the face similarity space, but the correlations of the similarity ratings were lower than in the single-category experiment, and the local arrangement of the faces showed differences. These differences between the behaviour of the model and data in the mixed-category experiment (figures 2 and 3) may reflect learning-dependent processes, such as biases in favour of the perception of a more familiar facial category (the ``other race'' effect). Our model demonstrates the utility of the LAC code for comparing the shape and texture of faces in an integrated fashion. That clearly cannot account for category-dependent biases in face processing, but higher-level models can build upon this measure of metric and texture differences.

We find it interesting that the human similarity space (figure 2) indicated a more biased perception than was exhibited by the LAC similarity space (figure 3). Note that neither the face comparison experiments by humans nor the LAC similarity measure took into account any labels or category information abut the facial images. The bias emerged from pairwise similarity judgements. It appears that the visual and social experience of the experimental subjects have influenced their judgements. The (American) viewers judged the EA faces as having greater within-category similarity than the EO faces. Moreover, both EA and EO faces were perceived as forming tight clusters on the 'race' axis by the humans (figure 2), but not the LAC measure, which distributed the faces more gradually (figure 3). In other words, the human subjects perceive similarities with with racial bias, but the LAC similarity measure does not exhibit this bias. It could be interesting to attempt to model the learning of bias in face perception to better understand such sociological/anthropological phenomena. It would be straightforward to train a classifier for such categories using the LAC similarity measure with a distribution of reference faces. Novel faces examined in low-dimensional similarity spaces like those of figures 2 and 3 may be categorized according to the proximity of reference faces. A related classification scheme has already been demonstrated in the context of multimedia technology by one of the authors (Lyons {\it et al.} 1998b). Prior exposure combined with social influence could model the development of racial and other biases in face perception. 

In the present work, each node in the facial graph received equal
weight in the similarity calculation. Previous studies have shown that
certain facial features play a more important role in facial
comparison than others (Bruce {\it et al.} 1987, 1993). Moreover, at
each node, it is quite likely that some spatial frequency and
orientation-selective filters carry more significant information than
others.  Therefore differential weighting of facial graph nodes and
filters within the similarity measure may further improve the
agreement with perceived facial similarity. We are currently
investigating methods for learning such differential weighting schemes
using discriminant analysis and neural network methods.

Figure 2 and 3 can be interpreted as depictions of aspects of the
``face space'' concept discussed extensively by Valentine (Valentine,
1991). It is interesting that the overall structure of the space
recovered by unsupervised analysis of the LAC similarity data shares
qualitative features with the ``face space'' as perceived by human
observers. However, as noted above details of the positioning of the
faces in this mixed-category space are not well captured by the model.

A notable feature of this work is that familiar facial categories emerged without supervised learning. Humans do not seem to need supervised training to classify faces and other objects.  In most cases, humans tend to discriminate faces and objects with surprisingly modest experience.  Some neurons in the inferior temporal cortex of infant monkeys (as young as six weeks old) exhibit selectivity for faces comparable to selectivity in adult monkeys (Rodman {\it et al.} 1991).  Human neonates in less than an hour after birth can imitate visually presented facial movements (Meltzoff and Moore 1983; Reissland 1988).  Twelve month-old infants seem to discriminate male faces from female faces (Leinbach and Fagot 1993). These findings suggest the possibility that the visual mechanism for face perception is present early in development and that face perception may take advantage of early maturing circuits. Our results are consistent with this view. 

\section*{Acknowledgements}
Michael Lyons is grateful to Christoph von der Malsburg for
stimulating discussions. Kaz Morikawa thanks Bela Julesz and the
Laboratory for Vision Research at Rutgers University for support.
Both authors thank Barlett Mel for the use of his laboratory in some
of the experiments, an anonymous referee for helping to clarify the
presentation through constructive criticism, and the Annenberg
Center at the University of Southern California for financial support
in the early stages of this project (grant 22-3801-1174,
P.I. M. J. Lyons)

\end{document}